\documentclass[conference]{IEEEtran}
\IEEEoverridecommandlockouts
\usepackage{cite}
\usepackage{amsmath,amssymb,amsfonts}
\usepackage{graphicx}
\usepackage{textcomp}
\usepackage{xcolor}
\usepackage{stfloats}
\usepackage{amsfonts}
\usepackage{bbm}
\usepackage{algorithm}
\usepackage{algpseudocode}
\usepackage{booktabs}
\usepackage{subfigure}
\usepackage{multirow}
\usepackage{graphicx}

\usepackage{url}
\def\BibTeX{{\rm B\kern-.05em{\sc i\kern-.025em b}\kern-.08em
T\kern-.1667em\lower.7ex\hbox{E}\kern-.125emX}}
\begin{document}

\title{Self-supervised Contrastive Learning for EEG-based Sleep Staging
\\
}

\author{\IEEEauthorblockN{1\textsuperscript{st} Xue Jiang}
\IEEEauthorblockA{\textit{Wuhan University} \\
Wuhan, China \\
jxt@whu.edu.cn}
\and
\IEEEauthorblockN{2\textsuperscript{nd} Jianhui Zhao 
\thanks{\IEEEauthorrefmark{1} Jianhui Zhao is corresponding author.}\IEEEauthorrefmark{1}}
\IEEEauthorblockA{\textit{Wuhan University} \\
Wuhan, China \\
jianhuizhao@whu.edu.cn}
\and
\IEEEauthorblockN{3\textsuperscript{rd}  Bo Du}
\IEEEauthorblockA{\textit{Wuhan University} \\
Wuhan, China \\
gunspace@163.com}
\and
\IEEEauthorblockN{4\textsuperscript{th} Zhiyong Yuan }
\IEEEauthorblockA{\textit{Wuhan University} \\
Wuhan, China \\
zhiyongyuan@whu.edu.cn}

}

\maketitle

\begin{abstract}
EEG signals are usually simple to obtain but expensive to label. Although supervised learning has been widely used in the field of EEG signal analysis, its generalization performance is limited by the amount of annotated data. Self-supervised learning (SSL), as a popular learning paradigm in computer vision (CV) and natural language processing (NLP), can employ unlabeled data to make up for the data shortage of supervised learning.
In this paper, we propose a self-supervised contrastive learning method of EEG signals for sleep stage classification. During the training process, we set up a pretext task for the network in order to match the right transformation pairs generated from EEG signals. In this way, the network improves the representation ability by learning the general features of EEG signals.
The robustness of the network also gets improved in dealing with diverse data, that is, extracting constant features from changing data. In detail, the network's performance depends on the choice of transformations and the amount of unlabeled data used in the training process of self-supervised learning.
Empirical evaluations on the Sleep-edf dataset demonstrate the competitive performance of our method on sleep staging (88.16\% accuracy and 81.96\% F1 score) and verify the effectiveness of SSL strategy for EEG signal analysis in limited labeled data regimes. All codes are provided publicly online.\footnote{ \url{https://github.com/XueJiang16/ssl-torch}}

\end{abstract}

\section{Introduction}
Electroencephalography (EEG) is a widely used neuroimaging technology in clinics, which is the measurement of the electric field generated by brain activity.
Precisely, EEG measures the potential difference produced from the electrical signals generated by the synaptic excitation of neurons to the scalp, which is generally about tens of $\mu$V \cite{Malhotra2014}.
Therefore, EEG reflects neurons' activity and can be used to study a wide range of brain processes, such as sleep monitoring, epilepsy detection, and so on. For example,
\textit{Ullah et al.}\cite{Ullah2018automated} take use of EEG signals to detect epileptic seizures.
\textit{Koushik et al.}\cite{koushik2018real} deploy a lightweight network on the mobile phone to achieve real-time sleep detection by analyzing the signal from the wearable EEG acquisition device.
\textit{Anumanchipalli et al.}\cite{anumanchipalli2019speech} use bi-LSTM to decode the EEG signals of epilepsy patients into speeches that humans can directly understand.
\textit{Hochberg et al.}\cite{hochberg2012reach} can control the robotic arm's actions with the human mind by using a brain-computer interface device.
Deep learning has achieved outstanding performance on these tasks, which is accomplished by the fully supervised learning with numerous manually labeled data.

However, the labeling of EEG signals is costly because the labeling of EEG signals, a tedious and time-consuming task, requires specific experts. For example, to label a segment of EEG signal with a duration of 24h, it takes about five hours of concentrated work by a well-trained expert \cite{Malhotra2014a}.
On the contrary, EEG signals are straightforward to obtain in clinical practice. A simple brain-computer interface can complete the acquisition of EEG signals, and as the acquisition time increasing, the device will acquire a large amount of data. Unfortunately, these data usually lack reliable annotations or are directly discarded, which actually have potential value for network training.

A novel approach that can take a large number of unlabeled data into the training process is Self-supervised learning (SSL)\cite{gidaris2018unsupervised,misra2016shuffle,mikolov2013distributed,devlin2018bert,chen2020simple}. As a result, the network can benefit from SSL with more external data.
EEG signals are electrical signals generated by neurons' activity, so they contain objective physical and physiological laws, which are exactly what SSL wants to mine.
Although SSL is popular and effective in other low labelled data regimes \cite{gidaris2018unsupervised,misra2016shuffle,mikolov2013distributed,devlin2018bert,chen2020simple}, few articles apply SSL into EEG processing.
The self-supervised learning tasks designed in these articles\cite{Saeed} \cite{banville2020uncovering} for EEG signals have limitations due to their reliance on prior knowledge of EEG signals, e.g., frequency domain information, rather than structurally general features. And the impacts of these SSL tasks on network representation are not deeply explored, resulting in limited self-supervised learning performance.

In this paper, we design a self-supervised contrastive learning method to conduct representation learning of EEG signals and apply it to sleep staging tasks.
Also, our proposed method makes more unlabeled data available for network training, thus surpassing the performances of the state-of-the-art models, reaching 88.163$\%$ accuracy on Sleep-edf dataset \cite{kemp2000analysis,goldberger2000components}.

Our contributions are summarized as follows.
\begin{itemize}
\item We design a self-supervised framework with contrastive learning for EEG signal representation learning.
Specifically, it measures the feature similarity of transformed signal pairs generated from EEG signals to learn the correlation between the signals.
Simultaneously, the influence of transformation compositions on the network representation ability is explored to obtain the optimal composition for the downstream task: sleep staging.
Self-supervised contrastive learning benefits more from a stronger transformation composition than supervised learning.
\item We apply the parameters learned by SSL to the downstream task: sleep staging, and design two kinds of experiments for network training: frozen backbone and fine-tuning.
The experimental results confirm the effectiveness of SSL for the downstream task and show that the performance of our method on Sleep-edf surpasses other fully supervised networks. In addition, the network uses the DOD-O and DOD-H datasets as additional data for SSL, and the classification accuracy on Sleep-edf has been improved in the subsequent downstream task, which shows that the increase in data volume has improved the ability of SSL to represent EEG signals.
\item Our experimental results illustrate that the proposed SSL method also performs well in limited sample learning. Based on SSL pretrained backbone, the network achieves 66.21$\%$ accuracy on Sleep-edf trained with only 10 samples per class (0.3$\%$ of the full training set), which reaches the level of fully supervised learning using 100 samples per class (3$\%$ of the full training set).
\end{itemize}

\section{Related work}
\subsection{EEG-based Sleep Staging}
Numerous studies \cite{Supratak2017,Tapia2020,Zhu2020,Chambon2019,Seo2020a,Perslev2019, Yildiz2009, Alickovic2018} have shown that using EEG signals to sleep staging is a quite common and reliable method. In the early work, researchers usually extract features from EEG signals by some traditional strategies (e.g., short-time Fourier transform (STFT) and discrete wavelet transform (DWT) \cite{Yildiz2009} ), and then fed them
into statistical classifiers (e.g., support vector machine (SVM) \cite{Alickovic2018}). In recent years, with the popularity of deep learning, a large number of methods based on CNN and RNN\cite{Supratak2017,Tapia2020,Zhu2020,Chambon2019,Seo2020a,Perslev2019} have emerged. For example, \textit{DeepSleepNet }\cite{Supratak2017} designs an end-to-end network combining Bi-LSTM and CNN to complete sleep classification tasks.
\textit{U-Time} \cite{Perslev2019} is inspired by U-net and design a one-dimensional network for EEG-based sleep staging on several datasets.
The deep learning-based models can automate the feature extraction process while having a better classification performance, thus becoming a better choice for sleep staging.
The common point of these above models is that they both belong to fully supervised learning network, which depends on the data and its corresponding labels.

\subsection{Self-supervised Learning}
Self-supervised learning is a machine learning paradigm that the network is trained using automatically generated labels instead of manually annotated labels.
The latest research in the field of machine learning and deep learning shows the potential of self-supervised models in learning generalized and robust representations \cite{gidaris2018unsupervised,misra2016shuffle,mikolov2013distributed,devlin2018bert,Sarkar,banville2020uncovering,Saeed}.
Therefore, SSL can learn from a large amount of unlabeled data to improve the representation ability of the network.

SSL has been widely used in many fields.
For example, in computer vision (CV), \cite{gidaris2018unsupervised} relies on the spatial structure of the image to artificially construct a self-supervised task for predicting the rotation angle of the image.
Based on the temporal structure of the video, a self-supervised task \cite{misra2016shuffle} is designed to make the network predict whether the video frame is shuffled. Similarly, in natural language processing (NLP), self-supervised models have a wide range of applications, such as the original model \textit{word2vec} \cite{mikolov2013distributed} and \textit{BERT} \cite{devlin2018bert}, which performs well on 11 NLP tasks.

It is worth mentioning that SSL is rarely applied to the field of biosignals, although it has the potential to make use of large amounts of unlabeled data. In \cite{Sarkar}, SSL is used as a way to extract features for ECG-based emotion recognition tasks. It designs a transformation recognition task for the network, that is, the original signal and multiple transformed signals are input into the network at the same time, and then the network needs to predict what kind of transformation the input has undergone.
In \cite{banville2020uncovering}, the author designs two pretext tasks: temporal context prediction and contrastive predictive coding to perform representation learning of EEG signals, and applies them to two downstream tasks: sleep staging and pathology detection. Also in \cite{Saeed}, SSL compares the time domain information and frequency domain information of signals for representation learning and finally deployed the model to IoT devices for different downstream applications.

\section{Method}\label{3}
\subsection{Self-supervised Contrastive Learning}
% \textbf{Notation:}

Inspired by \cite{chen2020simple}, we adopt a contrastive manner to design the self-supervised learning framework, which can learn representations of EEG signals by calculating the similarity in cosine metric space between different features. Note $s ({\boldsymbol{u},\boldsymbol{v}})$ as the similarity metric of $\boldsymbol{u},\boldsymbol{v}$:

\begin{equation}\label{e}
s(\boldsymbol{u},\boldsymbol{v})=cos(\theta_{\boldsymbol{u},\boldsymbol{v}}) = \frac{\boldsymbol{u}^T\boldsymbol{v}}{\left \| \boldsymbol{u} \right \|\left \|\boldsymbol{v} \right \|},s(\boldsymbol{u},\boldsymbol{v}) \in [0,1]
\end{equation}

When the cosine value $cos ( \theta_{\boldsymbol{u},\boldsymbol{v}})$ is close to 1 (i.e. $\theta_{\boldsymbol{u},\boldsymbol{v}} \rightarrow 0^{\circ}$), it indicates that the two vectors are more similar. When the cosine value $cos ( \theta_{\boldsymbol{u},\boldsymbol{v}})$ is close to 0 (i.e. $\theta_{\boldsymbol{u},\boldsymbol{v}}$ $\rightarrow$ $180^{\circ}$), it indicates that the two vectors are more different.

\begin{figure}[t]
\centering
\includegraphics[width=0.45\textwidth]{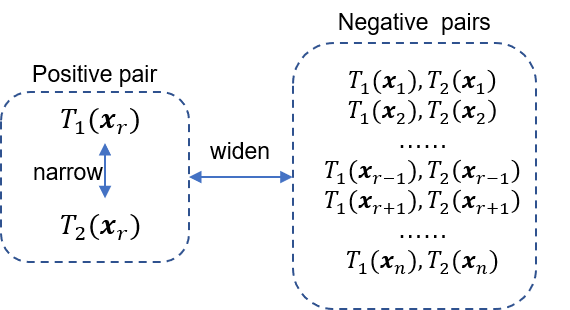}
\caption{Visual representations of proposed contrastive learning. $T_1(\boldsymbol{x}_1),T_2(\boldsymbol{x}_1),$ $\ldots, T_1(\boldsymbol{x}_n),T_2(\boldsymbol{x}_n)$ are transformed signal pairs from the orignal signals $\boldsymbol{x}_1, \boldsymbol{x}_2, \ldots, \boldsymbol{x}_n$.}
\label{trans}
\end{figure}

As shown in Fig.\ref{trans} and Algorithm 1, the original signals $\boldsymbol{x}_1, \boldsymbol{x}_2,\ldots,\boldsymbol{x}_n$ first undergo two different transformations $T_1(\cdot)$ and $T_2(\cdot)$ to generate $2n$ transformed signals $T_1(\boldsymbol{x}_1),T_1(\boldsymbol{x}_2),\ldots, T_1(\boldsymbol{x}_n),T_2(\boldsymbol{x}_1),T_2(\boldsymbol{x}_2),\ldots, T_2(\boldsymbol{x}_n)$, which then are sent to the network to extract features $\boldsymbol{t}_1,\boldsymbol{t}_2,\ldots,\boldsymbol{t}_n$. In the training process, for each $\boldsymbol{t}_i$, we require the network to find the homologous one in the remaining $2n-1$ samples by measuring feature similarity.
We take the homologous pair $T_1(\boldsymbol{x}_r),T_2(\boldsymbol{x}_r)$ as the positive pair while the others as negative pairs.
It is equivalent to a classification problem, so we use cross entropy to measure the loss of this task.
Note $\ell(i,j)$ as the contrast loss \cite{oord2018representation} of the pair $\left \{ \boldsymbol{t}_{i}, \boldsymbol{t}_{j}\right \}$,

\begin{equation}
\label{loss}
\ell(i,j)=-\log \frac{\exp \left(s\left(\theta_ { \boldsymbol{t}_{i}, \boldsymbol{t}_{j}}\right) / \tau\right)}{\sum_{k=1}^{2 N} \mathbbm{1}_{[k \neq i]} \exp \left(s(\theta_ { \boldsymbol{t}_{i}, \boldsymbol{t}_{k}}) / \tau\right)},
\end{equation}
where $\tau$ is a constant parameter. In our implementation, we configure $\tau = 0.5$ as default.
Then repeat the above process of taking positive pair for each pair to obtain multiple contrastive losses, and finally the loss fed back to the network is the average of all contrastive losses.
In this way, homologous pairs can become more similar, while the differences within heterologous pairs become larger. So the network is learning how to narrow the distance between the homologous transformation pairs. In essence, the network learns to extract the general features between the pairs, which is crucial to decode the transformed signals what they inherit from the original signals, thus improving the network's representation ability for the original signals.

\begin{algorithm}[t]
\label{alg}
\caption{Self-supervised contrastive learning algorithm}
\begin{algorithmic}[1]
\Require The orignal signals $\boldsymbol{x}_1, \boldsymbol{x}_2,\ldots,\boldsymbol{x}_n$, batchsize $N$, the network function $f$, transformation functions $T_1(\cdot),\ T_2(\cdot)$;
\Ensure	The loss $\mathcal{L}$ for the network;
\For{each sample $\boldsymbol{x}_i$ in a batchsize, $i \in \left \{ 1,2,\ldots,N \right \}$ }
\State make two signal transformations $T_1(\cdot),\ T_2(\cdot)$
\State $\boldsymbol{t}_i = f(T_1(\boldsymbol{x}_i))$
\State $\boldsymbol{t}_{i+N} = f(T_2(\boldsymbol{x}_i))$
\EndFor
\For {each $i,j \in \left \{ 1,2,\ldots,2N \right \}$}
\State $s(\theta_{\boldsymbol{t}_i,\boldsymbol{t}_j}) = \frac{{\boldsymbol{t}_i}^{T}\boldsymbol{t}_j}{\left \| \boldsymbol{t}_i \right \|\left \|\boldsymbol{t}_j \right \|}$
\EndFor
\State Calculate the contrast loss $\ell(i,j)$ as (\ref{loss});
\State The final loss $\mathcal{L} = \frac{1}{N} \sum_{i=1}^{N} \ell(i,i+N) $.
\end{algorithmic}
\end{algorithm}

\subsubsection{Pretext task for EEG: Signal Transformations}
\begin{figure}[htbp]
% \centering
\includegraphics[width=0.5\textwidth]{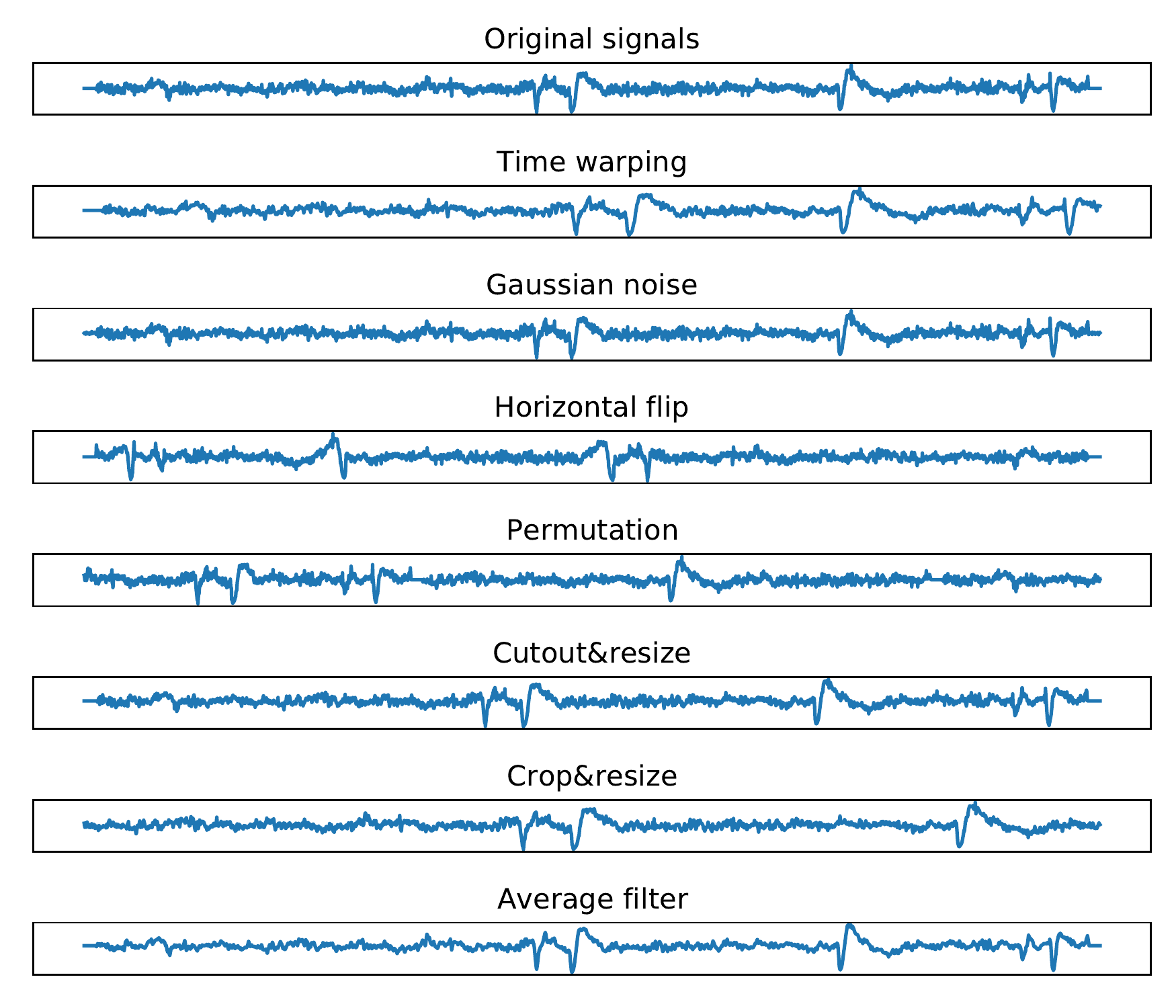}
\caption{Visual representations of signal transformations.}
\label{compare}
\end{figure}

The formulation as mentioned above requires the signal transformations $T(\cdot)$ to generate transformed signal pairs that enable the convolutional model to learn disentangled semantic representations useful for the following downstream task: sleep staging.
These transformations are described below and a sample of transformed signals is illustrated in Fig.\ref{compare}.

\textbf{Time warping}: Randomly selected segments of the original EEG signals are stretched or squeezed along the time axis. We segment original signals $S\left(t\right),\ t=1,2,\ldots,L$ into $\left\{S_1\left(t\right),S_2\left(t\right),\ldots,S_n\left(t\right)\right\}$. For each $S_i\left(t\right)$, we adopt transformations and get $S_i^\prime\left(t\right)=S_i\left(\omega t\right), \omega\in[0.25,4]$ is a random scale factor. Then, the transformed signals
$\left\{S_1^\prime\left(t\right),S_2^\prime\left(t\right),\ldots,S_n^\prime\left(t\right)\right\}$ are concatenated and resized to the original length $L$.

\textbf{Gaussian noise}: Random noise from a Gaussian distribution $N\left(t\right)=Gussian(\mu,\ \sigma^2)$ is added to the original EEG signals $S\left(t\right)$, which produces $ S^\prime\left(t\right)=S\left(t\right)+N\left(t\right)$.

\textbf{Horizontal flip}: The original EEG signals $S\left(t\right),\ t=1,2,\ldots,L$ are inversed in time axis as $S^\prime\left(t\right)=S\left(L-t+1\right)$.

\textbf{Permutation}: The original EEG signals $S\left(t\right)$ are randomly divided into several segments $\left\{S_1\left(t\right),S_2\left(t\right),\ldots,S_n\left(t\right)\right\}$ and shuffled into $\left\{S_{k_1}\left(t\right),S_{k_2}\left(t\right),\ldots,S_{k_n}\left(t\right)\right\}$, where $ \left\{k_1,\ k_2,\ldots,k_n\right\} $ is a permutation of $\left\{1,2,\ldots,n\right\}$, and then the shuffled segments are concatenated together.

\textbf{Cutout $\&$ resize}: The original EEG signals $S\left(t\right)$, $\ t=1,2,\ldots,L$ are randomly divided into several segments $\left\{S_1\left(t\right),S_2\left(t\right),\ldots,S_n\left(t\right)\right\} $ and we discard one $S_r\left(t\right)$ at random. Then the remaining segments $\left\{S_1\left(t\right),S_2\left(t\right),\ldots,S_{r-1}\left(t\right),S_{r+1}\left(t\right),\ldots,S_n\left(t\right)\right\}$ are concatenated together and resized to the original length $L$.

\textbf{Crop $\&$ resize}: The original EEG signals $S\left(t\right)$, $\ t=1,2,\ldots,L$ are randomly divided into several segments $\left\{S_1\left(t\right),S_2\left(t\right),\ldots,S_n\left(t\right)\right\}$ and we choose one
$S_r\left(t\right)$ at random. Then the chosen segment $S_r\left(t\right)$ are resized to the original length $L$.

\textbf{Average filter}: The original EEG signals $S\left(t\right)$ pass through the average filter with random filter length $k$ ranging from 3 to 10. The transformed signals are $S^\prime\left(t\right)=\frac{1}{k}\sum_{i=0}^{k-1}S\left(t+i\right)$.

\subsubsection{Network Architecture}We follow the model design proposed by \cite{Hannun2019} to build the backbone, which contains 18 1d convolutional layers with kernel sizes of 32. Moreover, the classifier is comprised of three fully-connected layers of 384, 192, 96 hidden units respectively and followed by a softmax output layer. We adopt ReLU\cite{nair2010rectified} as the activation unit for all the layers (except the output layer) and train a network on GPUs with SGD\cite{bottou2010large} optimizer with a momentum of 0.9. Dropout is used in each residual block with a rate of 0.2, and L2 regularization is applied with a rate of 0.0001.
Batchsize is set as 512 in SSL and 256 in the classification task.
The networks are trained for 70 epochs in SSL and classification task respectively, and both adopt cosine warmup strategy \cite{loshchilov2016sgdr}.

\subsection{Downstream Task: Sleep Staging}
\begin{figure*}[t]
% \centering
\includegraphics[width=\textwidth]{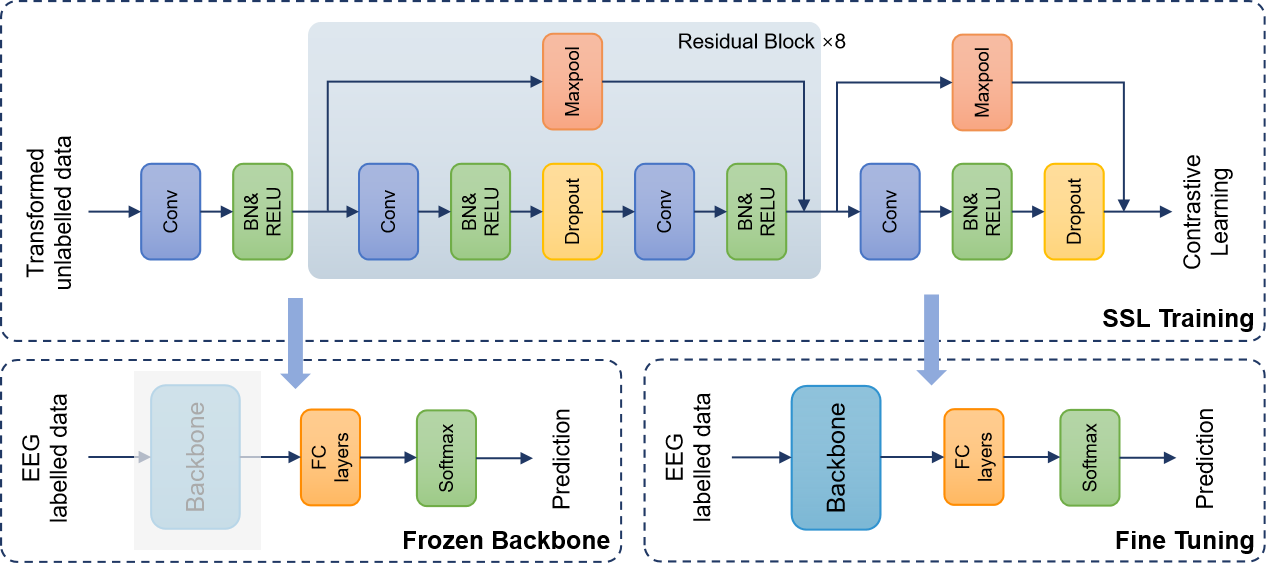}
\caption{The network architecture of self-supervised contrastive learning. First, the model is trained with unlabeled data to learn EEG representations. Then, the weights are transferred to two sleep staging networks, one freezes the backbone and only trains the classifier, while the other directly uses the weights as a pre-training model for fine-tuning.}
\label{net}
\end{figure*}
Sleep staging, which is representative of current challenges in machine learning-based analysis of EEG, is typically regarded as a five-class classification problem where the possible predictions are W (Wake), REM (Rapid Eye Movement periods), N1, N2 and N3 (Non-REM).
We apply models trained by EEG-based SSL on sleep staging and design two kinds of experiments to evaluate the performance of the network on the sleep staging task.

\subsubsection{Classifier training with frozen backbone}
To evaluate the learned representations, we follow the widely used linear evaluation manner \cite{oord2018representation,zhang2016colorful,bachman2019learning,kolesnikov2019revisiting}.
% (Zhang et al., 2016; Oord et al., 2018; Bachman et al., 2019; Kolesnikov et al., 2019)
We load and freeze the SSL-trained backbone for the classification task and train the classifier with the frozen backbone, as shown in Fig.\ref{net}. This evaluation protocol is conducted to check if the model learns the discriminative features for EEG representation, and the test accuracy is used as a proxy to evaluate the signal representation quality of SSL. The specific experiments and results are in section \ref{seca}.
\subsubsection{Fully supervised fine tuning}
The backbone generated by SSL is used as a pretrained model, and fine-tuning is performed on the basis of these parameters. This can evaluate the improvement degree of SSL for sleep staging. The specific experiments and results are in section \ref{seca}.

\section{Experiments}
\subsection{Data Preparation}
\begin{table}[t]
\caption{Detailed information about sleep databases used in this study. $f_s$ means signal sampling rate.}
\centering
\renewcommand\arraystretch{1.25}
\label{dataset}
\setlength{\tabcolsep}{1mm}{
\begin{tabular}{@{}ccccc@{}}
\toprule
& DOD-H & DOD-O & Sleep-edf & Sleep-edfx \\ \midrule
Subjects & 25 & 55 & 20 & 100 \\
Attribute & Health & Sleep disorder & Health & Health \& Sleep disorder \\
Age & 35.32±7.51 & 45.6±16.5 & 28.74±1.73 & 54.70±4.74 \\
Records & 24665 & 54197 & 20667 & 238989 \\
$f_s$ (Hz) & 250 & 250 & 100 & 100 \\
Wake (\%) & 13.4 & 20.0 & 23.7 & 29.8 \\
N1 (\%) & 7.10 & 6.11 & 6.23 & 10.5 \\
N2 (\%) & 46.7 & 46.8 & 41.3 & 37.2 \\
N3 (\%) & 14.4 & 11.9 & 12.4 & 8.14 \\
REM (\%) & 18.2 & 14.8 & 16.3 & 14.3 \\ \bottomrule
\end{tabular}}
\end{table}
% \subsubsection{Datasets}
We use four publicly available datasets to evaluate the performance of proposed methods in section \ref{3}. The details of the datasets are summarized in Table \ref{dataset} and a brief description of each dataset is provided below.

\textit{Sleep-edf} \cite{kemp2000analysis,goldberger2000components}. It consists of 20 healthy subjects whose ages range from 25 years old to 34 years old. The sample rate is 100 Hz and there are two channels available for EEG signals: Fpz-Cz and Pz-Oz.

\textit{Sleep-edfx} \cite{kemp2000analysis,goldberger2000components}. It is the extended version of Sleep-edf and contains 197 whole-night PSG sleep recordings.
The dataset can be divided into two types of files: the 153 SC files were obtained in 78 healthy subjects aged 25-101 without any sleep-related medication, and the 44 ST files were obtained in 22 subjects who had mild difficulty falling asleep.
The sample rate is 100 Hz and there are two channels available for EEG signals: Fpz-Cz and Pz-Oz.

\textit{Dreem Open Dataset - Obstructive (DOD-O)} \cite{guillot2020dreem}. The dataset consists of PSG recordings from 55 patients suffering from obstructive sleep apnea (OSA). EEG signals in the dataset are composed of 8 EEG derivations (C3-M2, C4-M1, F3-F4, F3-M2, F4-O2, F3-O1, O1-M2, O2-M1) sampled at 250 Hz.

\textit{Dreem Open Dataset - Healthy (DOD-H)} \cite{guillot2020dreem}.
The dataset consists of PSG recordings from 25 healthy sleepers without sleep disorders between the ages of 18 and 65. EEG signals in the dataset are composed of 12 EEG derivations (C3-M2, F4-M1, F3-F4, F3-M2, F4-O2, F3-O1, FP1-F3, FP1-M2, FP1-O1, FP2-F4, FP2-M1, FP2-O2) sampled at 250 Hz.

The sampling rate and subjects of these datasets are different, and the experiment needs to mix the data of multiple datasets for training, so we uniformly process all the data into sequences with the length of 3072. In addition, the normalization transformation with mean of 0 and variance of 0.5 is applied to these data to ensure the input distribution consistency. In this way, the network can be avoided from being disturbed by the uneven data distribution.

% \subsubsection{Preprocessing}

\subsection{Results}\label{seca}
% \usepackage{booktabs}
% \usepackage{multirow}
% \usepackage{graphicx}
\begin{table*}[t]
\centering
\caption{The results of study in data amout and transformation compositions on Sleep-edf dataset. Baseline represents a fully supervised network. $\uparrow$ means the performance is better than the baseline while $\downarrow$ is the opposite meaning and -- means the performance remains the same.}
\label{vs}
\resizebox{\textwidth}{!}{%
\begin{tabular}{@{}ccccccccc@{}}
\toprule
\multirow{2}{*}{SSL data} & \multirow{2}{*}{Transformation composition} & \multicolumn{2}{c}{Overall Metrics} & \multicolumn{5}{c}{Per Class F1(\%)} \\ \cmidrule(l){3-4} \cmidrule(l){5-9}
& & Acc(\%) & F1(\%) & W & N1 & N2 & N3 & REM \\ \cmidrule(r){1-2} \cmidrule(l){3-4} \cmidrule(l){5-9}
Baseline (no SSL) & None & 86.60 & 79.51 & 93.27 & 45.21 & 89.60 & 86.13 & 83.36 \\ \cmidrule(r){1-2} \cmidrule(l){3-4} \cmidrule(l){5-9}
\multirow{2}{*}{Sleep-edf} & Crop\&resize\,+\,Permutation & 86.58 $\downarrow$ & 77.40 $\downarrow$ & 93.32 $\uparrow$ & 34.71 $\downarrow$ & 89.82 $\uparrow$ & 86.46 $\uparrow$ & 82.70 $\downarrow$ \\
& Crop\&resize\,+\,Crop\&resize & 85.97 $\downarrow$ & 78.98 $\downarrow$ & 92.48 $\downarrow$ & 44.87 $\downarrow$ & 89.04 $\downarrow$ & 87.01 $\uparrow$ & 81.50 $\downarrow$ \\ \cmidrule(r){1-2} \cmidrule(l){3-4} \cmidrule(l){5-9}
\multirow{2}{*}{Sleep-edfx} & Crop\&resize\,+\,Permutation & 88.13 $\uparrow$ & 82.64 $\uparrow$ & 93.54 $\uparrow$ & 53.85 $\uparrow$ & 91.31 $\uparrow$ & 88.16 $\uparrow$ & 86.34 $\uparrow$ \\
& Crop\&resize\,+\,Crop\&resize & 87.03 $\uparrow$ & 80.39 $\uparrow$ & 93.07 $\downarrow$ & 47.57 $\uparrow$ & 89.60 \,-- & 86.68 $\uparrow$ & 85.01 $\uparrow$ \\ \cmidrule(r){1-2} \cmidrule(l){3-4} \cmidrule(l){5-9}
\multirow{2}{*}{Dod-O\,+\,Dod-H\,+\,Sleep-edf} & Crop\&resize\,+\,Time warping & 87.78 $\uparrow$ & 82.12 $\uparrow$ & 93.48 $\uparrow$ & 53.55 $\uparrow$ & 90.20 $\uparrow$ & 87.35 $\uparrow$ & 86.01 $\uparrow$ \\
& Crop\&resize\,+\,Permutation & 87.71 $\uparrow$ & 80.94 $\uparrow$ & 94.39 $\uparrow$ & 48.94 $\uparrow$ & 90.27 $\uparrow$ & 86.28 $\uparrow$ & 84.81 $\uparrow$ \\ \cmidrule(r){1-2} \cmidrule(l){3-4} \cmidrule(l){5-9}
\multirow{2}{*}{Dod-O\,+\,Dod-H\,+\,Sleep-edfx} & Crop\&resize\,+\,Time warping & 87.39 $\uparrow$ & 81.49 $\uparrow$ & 94.00 $\uparrow$ & 51.84 $\uparrow$ & 89.53 $\downarrow$ & 88.45 $\uparrow$ & 83.63 $\uparrow$ \\
& Crop\&resize\,+\,Permutation & 88.16 $\uparrow$ & 81.96 $\uparrow$ & 93.85 $\uparrow$ & 50.35 $\uparrow$ & 90.81 $\uparrow$ & 88.39 $\uparrow$ & 86.39 $\uparrow$ \\ \bottomrule
\end{tabular}%
}
\end{table*}
\subsubsection{Study on transformation pairs}

\begin{figure}[t]
% \centering
\includegraphics[width=0.5\textwidth]{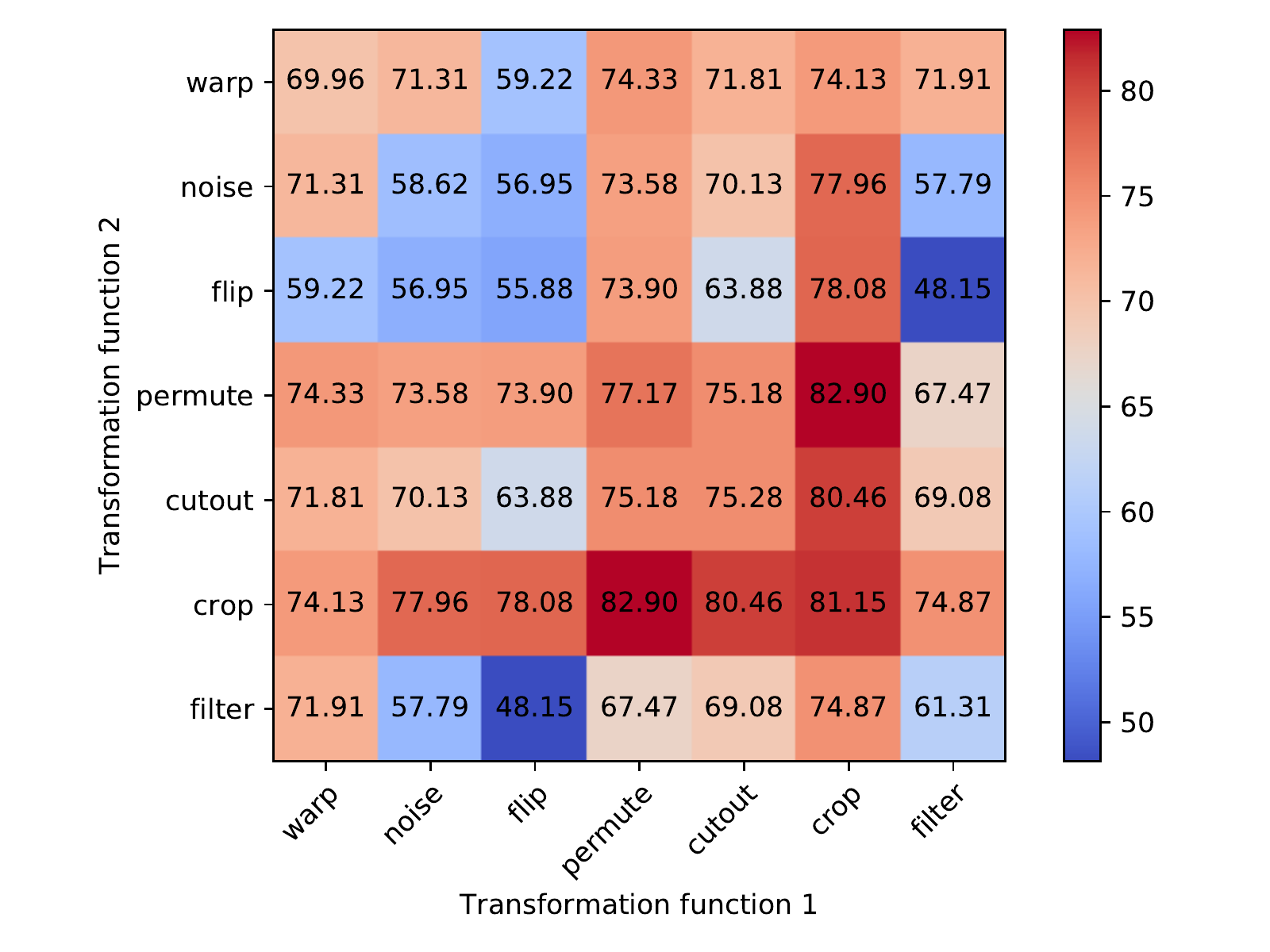}
\caption{Comparison of transformation compositions.}
\label{pairs}
\end{figure}

In the process of self-supervised training, we notice that different transformation compositions have a significant impact on network performance.
To further explore the effects of transformation composition, we investigate the performance of our network when applying different types of transformation pairs on Sleep-edf.

And then we train the classifier with the frozen backbone on Sleep-edf and obtain the test accuracy of sleep staging as shown in Fig.\ref{pairs},
which is used as a proxy to evaluate the signal representation quality of SSL with different transformation pairs.
The composition of \textit{Permutation} and \textit{Crop\&resize} stands out with the highest accuracy 82.90\%.

\subsubsection{Self-supervised contrastive learning}

In order to further explore the influence of different transformation pairs on the downstream task, we choose the models learned by SSL with different transformation pairs as pre-training, and then perform fine-tuning on the Sleep-edf dataset. The results are shown in Table \ref{vs}. It highlights that the composition of \textit{Permutation} and \textit{Crop\&resize} can get better results in most cases.

In addition, we select different amounts of unlabeled data for SSL training, and the influence of fine-tuned performance on downstream tasks is shown in Table \ref{vs}. It illustrates that the more unlabeled data SSL is fed, the better performance the network has.

The best results are 88.16\% accuracy and 81.96\% F1 score. Compared with Baseline, which takes random parameters as initialization, our method gets two percents increase in accuracy and F1 score.

\subsubsection{SSL versus Supervised learning}

We compare our methods with the state-of-the-art supervised learning models in performance on Sleep-edf and Sleep-edfx datasets and the results are shown in Table \ref{su}.
Our method stands out on Sleep-edf dataset in both overall metrics and the per class F1 score. Moreover, our approach currently has an obvious improvement on the Sleep-edfx dataset despite it only outperforms other model (U-Time) in two classes (W and N2).
The possible reason for this phenomenon is that the EEG signals from Sleep-edfx dataset have more complicated distributions, which may be caused by the physiological bias between the elderly and the young, the healthy and the unhealthy.

\begin{table*}[t]
\centering
\caption{The comparison of the performance with the state-of-the-art methods on Sleep-edf and Sleep-edfx datasets. SC means the healthy subjects while ST means subjects with sleep disorders. The bold numbers represents the highest performances. }
\label{su}
\resizebox{\textwidth}{!}{%
\begin{tabular}{@{}cccccccccc@{}}
\toprule
\multirow{2}{*}{Method} &
\multirow{2}{*}{Dataset} &
\multirow{2}{*}{Subjects} &
\multicolumn{2}{c}{Overall Metrics} &
\multicolumn{5}{c}{Per Class F1(\%)} \\ \cmidrule(l){4-5}\cmidrule(l){6-10}
& & & Acc(\%) & F1(\%) & W & N1 & N2 & N3 & REM \\ \cmidrule(r){1-3}\cmidrule(l){4-5}\cmidrule(l){6-10}
DeepSleepNet\cite{Supratak2017} & Sleep-edf & 20\,SC & 82.00 & 76.88 & 84.70 & 46.60 & 85.90 & 84.80 & 82.40 \\
MultitaskSleepNet\cite{phan2018joint} & Sleep-edf & 20\,\,SC & 82.25 & 74.72 & 77.32 & 40.54 & 87.45 & 85.99 & 82.31 \\
SleepEEGNet\cite{mousavi2019sleepeegnet} & Sleep-edf & 20\,SC & 84.26 & 79.66 & 89.19 & 52.19 & 86.77 & 85.13 & 85.02 \\
IITNET \cite{Seo2020} & Sleep-edf & 20\,SC & 83.60 & 76.54 & 87.10 & 39.20 & 87.80 & 87.70 & 80.90 \\
\textit{Ours(Baseline) } & Sleep-edf & 20\,SC & 86.60 & 79.51 & 93.27 & 45.21 & 89.60 & 86.13 & 83.36 \\
\textit{Ours(SSL) }&
Sleep-edf &
20\,SC &
\textbf{88.16} &
\textbf{81.96} &
\textbf{93.85} &
\textbf{50.35} &
\textbf{90.81} &
\textbf{88.39} &
\textbf{86.39} \\ \midrule
SleepEEGNet\cite{mousavi2019sleepeegnet} & Sleep-edfx & 78\,SC\,+\,22\,ST & 80.03 & 73.55 & 91.72 & 44.05 & 82.49 & 73.45 & 76.06 \\
U-Time\cite{Perslev2019} & Sleep-edfx & 78\,SC & 81.30 & 76.26 & 92.03 & 51.03 & 83.45 & 74.56 & 80.23 \\
U-Time\cite{Perslev2019} &
Sleep-edfx &
22\,ST &
83.16 &
78.61 &
87.14 &
\textbf{51.51} &
86.44 &
\textbf{84.24} &
\textbf{83.70} \\
\textit{Ours(Baseline)} & Sleep-edfx & 78\,SC\,+\,22\,ST & 82.07 & 75.32 & 92.69 & 42.44 & 85.25 & 77.89 & 78.32 \\
\textit{Ours(SSL)} &
Sleep-edfx &
78\,SC\,+\,23\,ST &
\textbf{84.42} &
\textbf{78.95} &
\textbf{93.65} &
49.26 &
\textbf{87.26} &
83.41 &
81.19 \\ \bottomrule
\end{tabular}%
}
\end{table*}

\begin{figure*}[t!]
\centering
\subfigure[Performance in accuracy]{\includegraphics[width=0.48\linewidth]{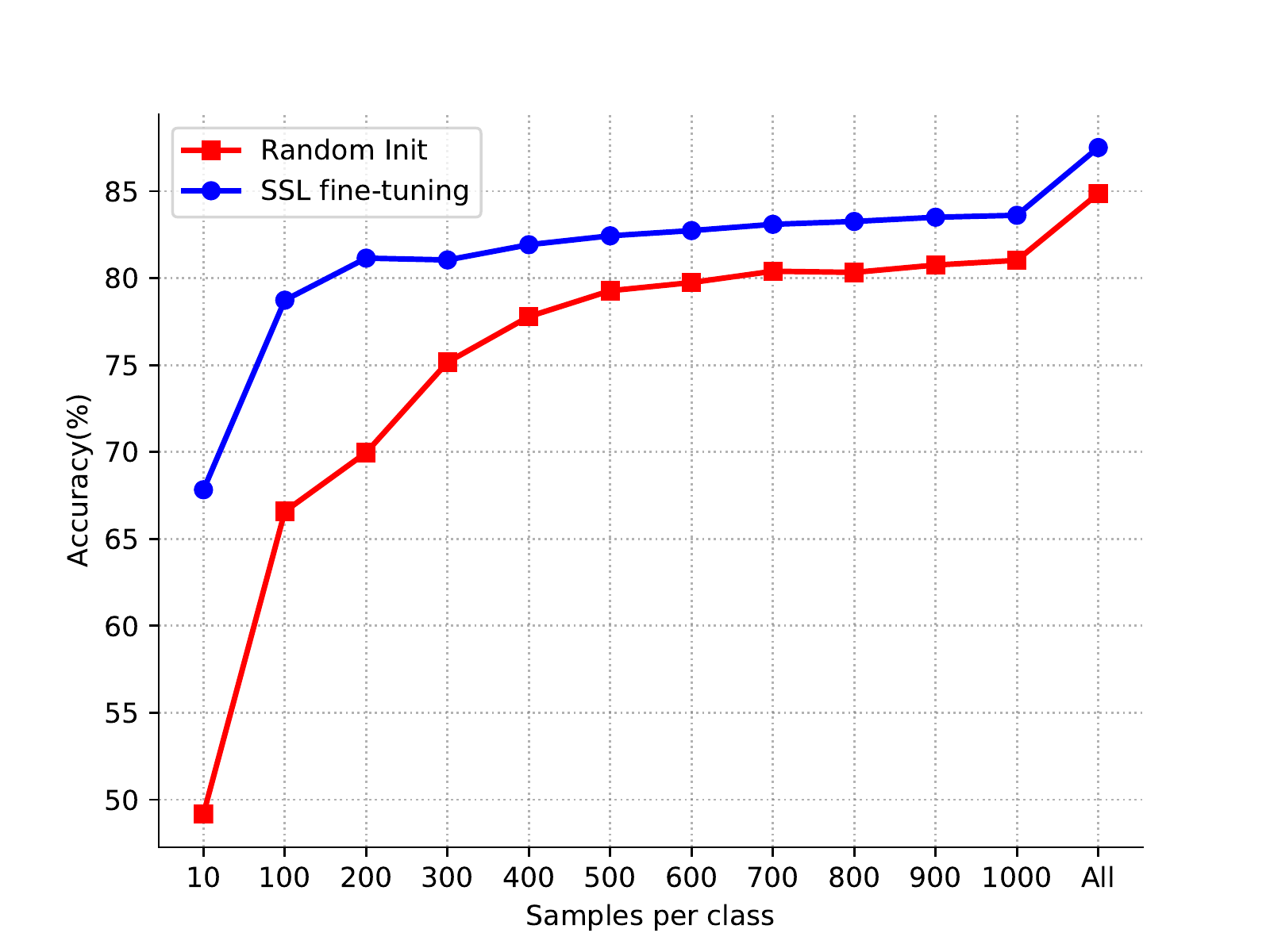}}
\subfigure[Performance in F1 score]{\includegraphics[width=0.48\linewidth]{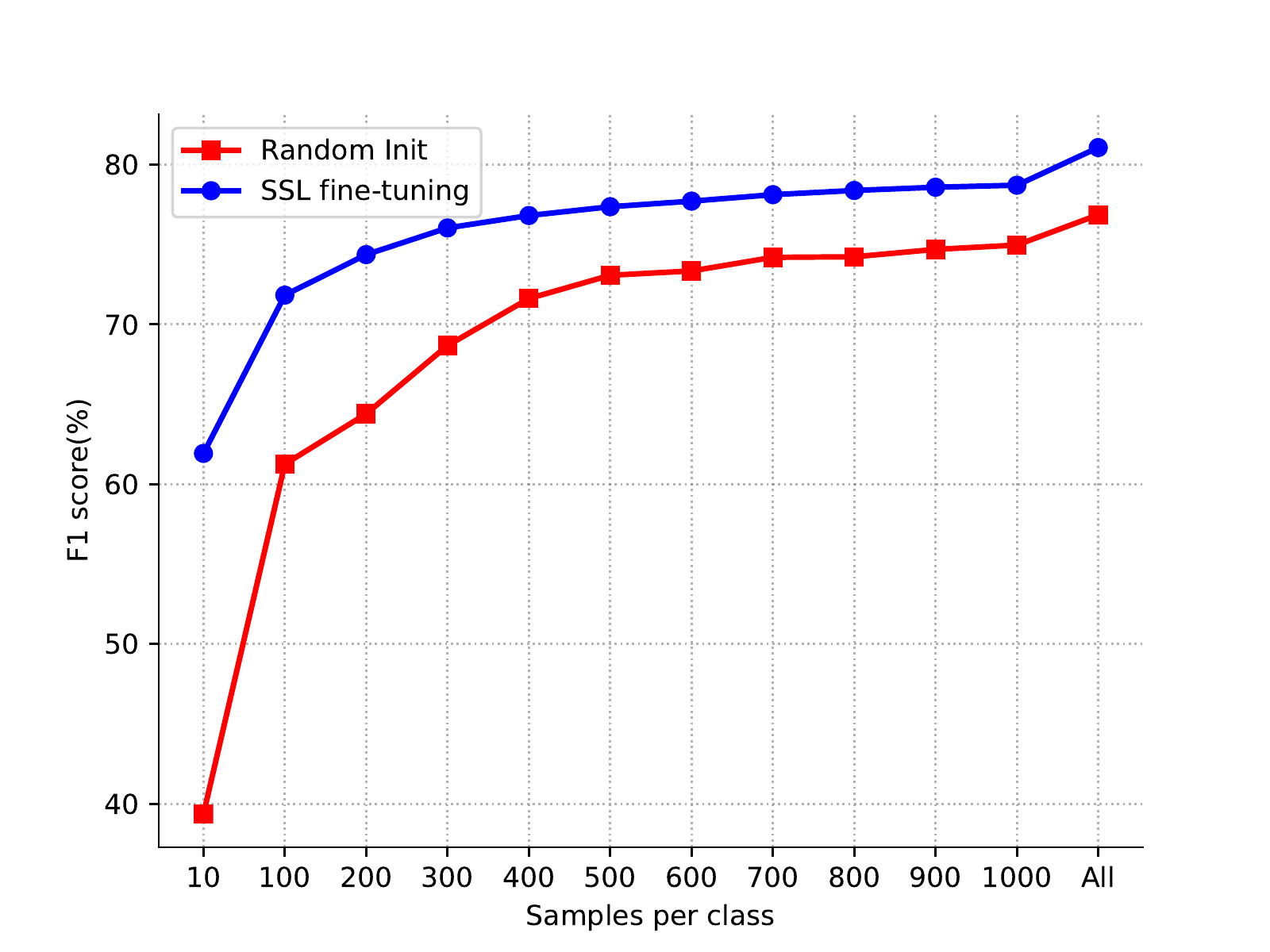}}
\caption{Impact of number of labeled samples per class on doenstream performance. Feature extractors are trained with our method and a fully supervised learning. `All` means all available training samples are used. More available training samples lead to a better performance while SSL model achieves a much higher performance than a fully supervised model when limited labeled samples are available.}
\label{line}
\end{figure*}

\begin{table}[t]
\centering
\caption{Results of number of labeled samples per class on doenstream performance. All the numbers are the average of 100 independent repetitions.}
\label{limited}
\begin{tabular}{@{}ccccc@{}}
\toprule
\multirow{2}{*}{\begin{tabular}[c]{@{}c@{}}Samples\\ per\\ class\end{tabular}} & \multicolumn{2}{c}{Random Initialization} & \multicolumn{2}{c}{SSL Fine-tuning} \\ \cmidrule(l){2-5}
& Acc(\%) & F1(\%) & Acc(\%) & F1(\%) \\ \midrule
10 & 49.176±4.025 & 39.367±6.863 & 67.826±5.639 & 61.925±5.468 \\
100 & 66.580±2.646 & 61.254±2.429 & 78.731±0.894 & 71.827±1.225 \\
200 & 69.966±2.258 & 64.401±2.214 & 81.149±0.632 & 74.363±1.012 \\
300 & 75.163±2.627 & 68.674±2.775 & 81.045±0.707 & 76.034±0.775 \\
400 & 77.786±1.789 & 71.613±1.817 & 81.922±0.632 & 76.809±0.775 \\
500 & 79.275±0.949 & 73.070±1.183 & 82.434±0.548 & 77.359±0.707 \\
600 & 79.745±1.342 & 73.342±1.732 & 82.733±0.447 & 77.708±0.707 \\
700 & 80.396±1.012 & 74.192±1.612 & 83.094±0.548 & 78.114±0.632 \\
800 & 80.329±1.789 & 74.227±2.145 & 83.259±0.548 & 78.376±0.632 \\
900 & 80.752±0.894 & 74.690±1.225 & 83.502±0.548 & 78.579±0.548 \\
1000 & 81.024±0.894 & 74.961±1.449 & 83.614±0.548 & 78.705±0.707 \\
All & 84.864±0.316 & 76.857±1.225 & 87.510±0.316 & 81.061±0.632 \\ \bottomrule
\end{tabular}
\end{table}

\begin{figure*}[t!]
\centering
\subfigure[tSNE visiualization of SSL features on Sleep-edf dataset.]{\includegraphics[width=0.48\linewidth]{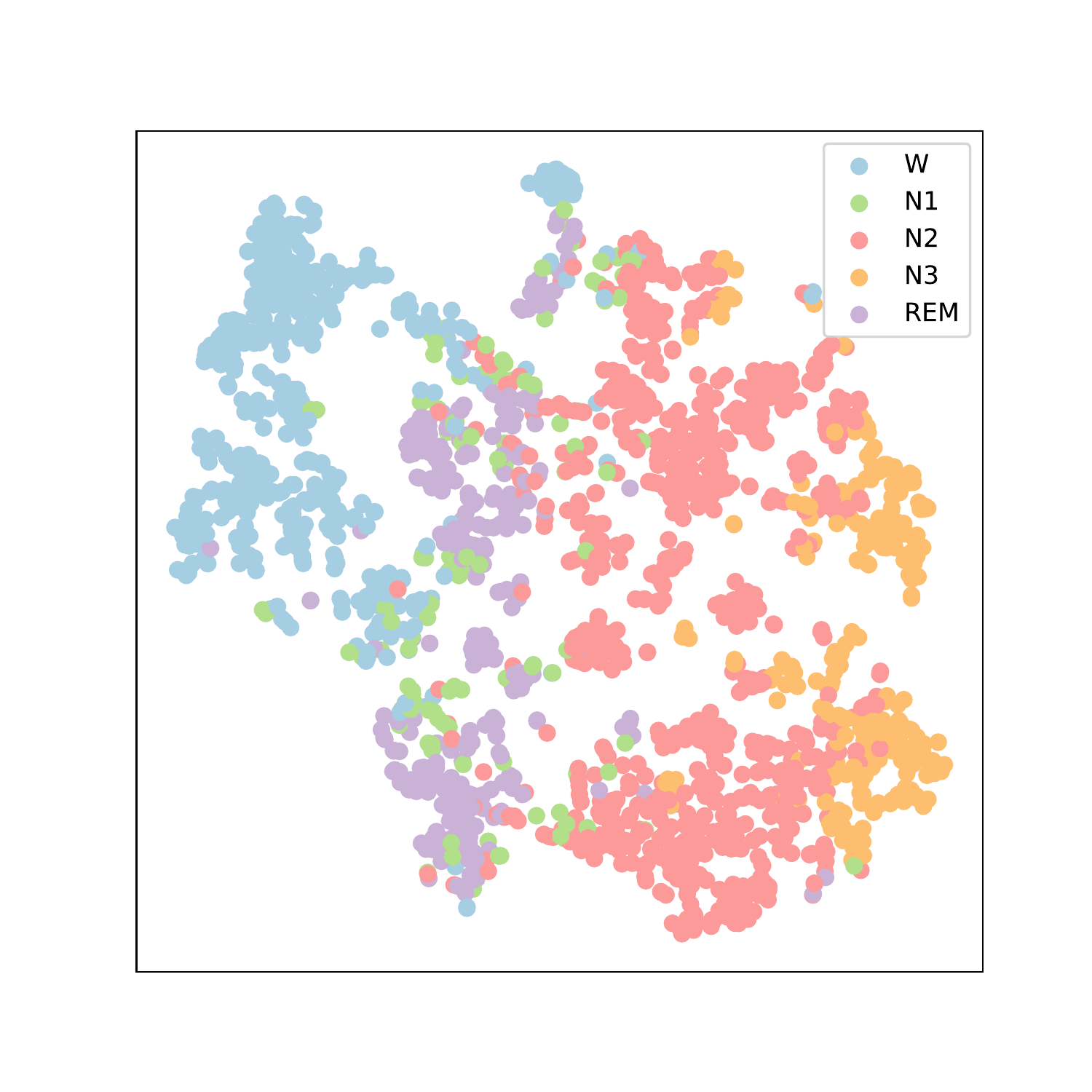}}
\subfigure[tSNE visiualization of fine-tuned features on Sleep-edf dataset. ]{\includegraphics[width=0.48\linewidth]{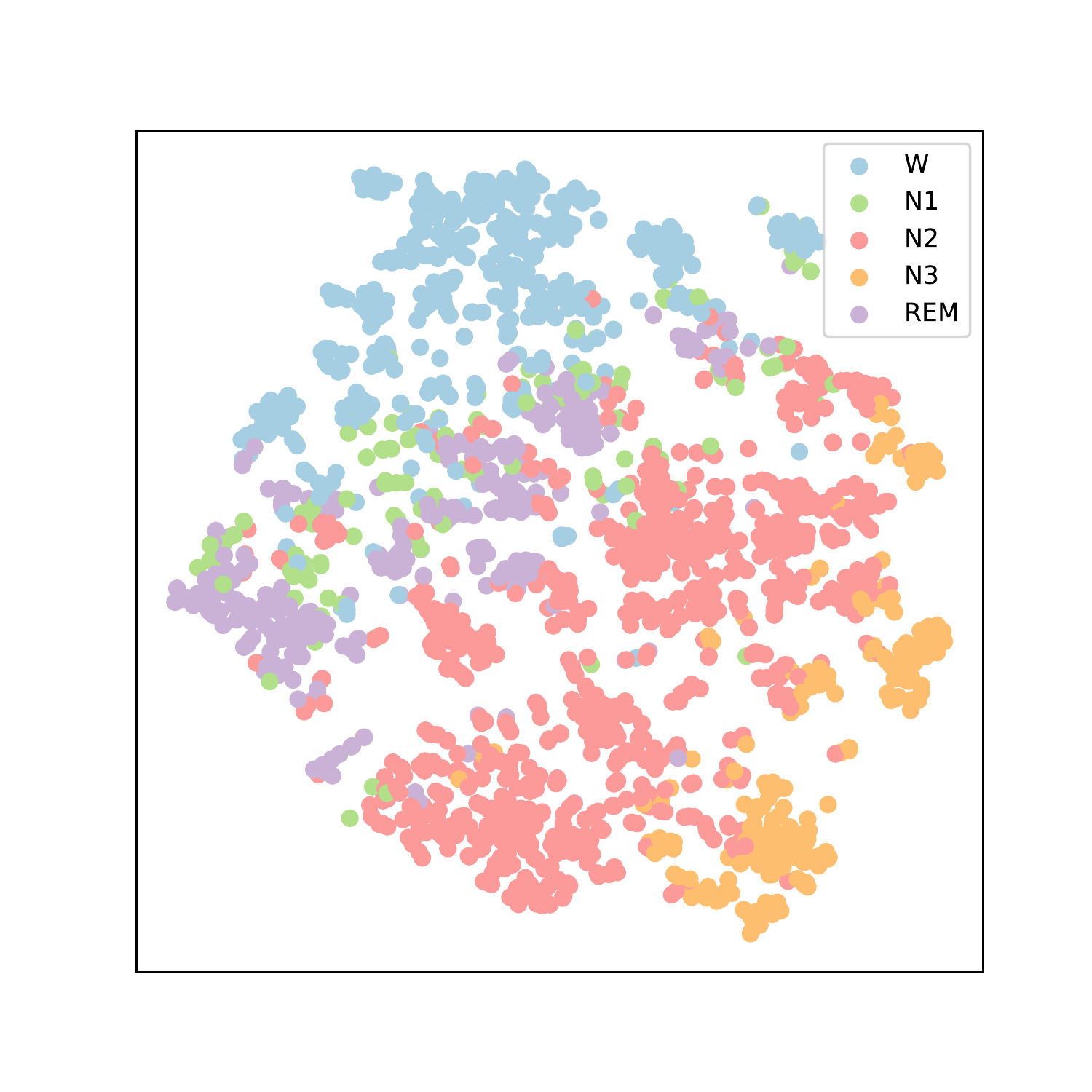}}
\caption{tSNE visiualizations. The scatterplot shows the feature distribution of the five sleep stages. Each point represents the features extracted from a 30s window EEG signals and the colors represents five kinds of sleep stages. SSL obtains a discriminative signal representation despite the fact that no labels are available during training, and the unsupervised clustering effect of the features is competitive enough with the effect of supervised learning.}
\label{0}
\end{figure*}

\subsubsection{Limited labeled sample learning}

In order to explore the effectiveness of the proposed method on limited labeled samples, we use unlabeled data from the Dod-O, Dod-H and Sleep-edfx datasets to pretrain the self-supervised contrast network for downstream tasks on Sleep-edf. We compare the performance with a standard supervised network trained only on certain labeled samples of Sleep-edf, as shown in Table \ref{limited}.

Specifically, we use different amounts of labeled samples per class and the full training set for fine-tuned process and the test set with about 4000 samples remains the same. Table \ref{limited} shows the average F1 score and average accuracy of 100 independent repetitions, where different signals are sampled to train the model in each run. Fig.\ref{line} (a)(b) are visualizations of Table \ref{limited}. In all cases, even with limited labeled data, the self-supervised model performs better than the supervised network. It is obvious that
self-supervised contrastive learning effectively extracts generalized features from unlabeled data.

\subsubsection{Feature visualization}
Although the features learned through SSL yield competitive performance on sleep staging, it is not clear which structure SSL captures and how the feature distribution changes after fine-tuning. In order to explore this, we use t-Distributed Stochastic Neighbor Embedding (tSNE) to visualize the 2000-dimensional embeddings obtained on Sleep-edf, and use the models learned by SSL and fine-tuning.

In the embedding of Sleep-edf learned through SSL and fine-tuning, the distribution following different sleep stages can be noticed in Fig.\ref{0}. Upon inspection of the distribution of samples at various stages, an exact group emerged. It can be observed that the distribution of the five stages is concentrated and distinguishable and especially N1 overlaps most with REM.

Comparing the two figures, it can be found that SSL has the ability to learn distinguished features for sleep staging without any access to the true labels, which indicates that self-supervised contrast learning can make the network sensitive to deep differences of data.
Besides, it is noticeable that the confusion between N1 and REM is an important factor hindering the improvement of the network performance.

% \begin{figure}[t]
% \centering
% \subfigure[tsne]{\includegraphics[width=0.45\linewidth]{tsne.png}}
% \subfigure[tsne finetune]{\includegraphics[width=0.45\linewidth]{tsne_c.png}}
% \caption{Comparison of (a)ground truth, (b)raw predictions, (c)processed predictions of 450-minute signals shows that the output jitters decrease a lot after the Markov-based sequential correction algorithm. Each grid in the figure represents the stage of each sample, and the five states are distinguished by color shade. } \label{compare}
% \end{figure}
\section{Conclusion}
In this work, we design a framework of self-supervised learning to improve the performance of sleep staging by enhancing the network representation ability of EEG signals. The pretext task in SSL traning is to match the correspondent transformation signal pairs among all the transformation signal pairs generated from EEG signals.

The experimental results demonstrate that: 1) our SSL method outperforms the state-of-the-art methods on sleep staging. 2) more unlabeled data available for SSL training can improve the representation of the network. 3) the representation learned by SSL also encodes physiologically information for sleep staging, showing their potential to uncover meaningful general features in unlabeled data. 4) even in limited labeled sample learning, our method still maintains an exceptional performance.

Although our method is proposed for EEG-based sleep staging, it is easy to generalize to other time-series signals and other applications related to EEG signals. We will further explore how to establish multi-task models on other datasets at a small cost.

% \newpage
% \section*{References}

% \bibitem{b1} G. Eason, B. Noble, and I. N. Sneddon, ``On certain integrals of Lipschitz-Hankel type involving products of Bessel functions,'' Phil. Trans. Roy. Soc. London, vol. A247, pp. 529--551, April 1955.
% \bibitem{b2} J. Clerk Maxwell, A Treatise on Electricity and Magnetism, 3rd ed., vol. 2. Oxford: Clarendon, 1892, pp.68--73.
% \bibitem{b3} I. S. Jacobs and C. P. Bean, ``Fine particles, thin films and exchange anisotropy,'' in Magnetism, vol. III, G. T. Rado and H. Suhl, Eds. New York: Academic, 1963, pp. 271--350.
% \bibitem{b4} K. Elissa, ``Title of paper if known,'' unpublished.
% \bibitem{b5} R. Nicole, ``Title of paper with only first word capitalized,'' J. Name Stand. Abbrev., in press.
% \bibitem{b6} Y. Yorozu, M. Hirano, K. Oka, and Y. Tagawa, ``Electron spectroscopy studies on magneto-optical media and plastic substrate interface,'' IEEE Transl. J. Magn. Japan, vol. 2, pp. 740--741, August 1987 [Digests 9th Annual Conf. Magnetics Japan, p. 301, 1982].
% \bibitem{b7} M. Young, The Technical Writer's Handbook. Mill Valley, CA: University Science, 1989.
% \end{thebibliography}
\bibliographystyle{ieeetr}
\bibliography{ref}

\end{document}